\documentclass{ifacconf}

\usepackage{inputenc}

\usepackage{graphicx}      
\usepackage{natbib}        
\usepackage{multirow}

\usepackage{amsmath}        
\usepackage{amssymb}        
\usepackage{algorithm}      
\usepackage{algpseudocode}  

\usepackage{xspace}

\usepackage{url}

\algrenewcommand\algorithmicrequire{\textbf{Input:}}
\algrenewcommand\algorithmicensure{\textbf{Output:}}
\newcommand{\CommentState}[1]{\Statex\hspace{\algorithmicindent}{\color{blue}// #1}}

\newcommand{\norm}[1]{\left\lVert #1 \right\rVert}

\newcommand{\Ne}{{N_\mathrm{e}}}
 
\DeclareMathOperator*{\argmin}{\arg\!\min}

\newcommand{\x}{\boldsymbol{x}}
\newcommand{\z}{\boldsymbol{z}}

\newtheorem{definition}{Definition}
\newtheorem{assumption}{Assumption}
\newtheorem{proposition}{Proposition}
\newtheorem{remark}{Remark}

\usepackage[table]{xcolor} 
\usepackage{makecell}
\usepackage{array} 

\makeatletter
\let\old@ssect\@ssect 
\makeatother

\usepackage{hyperref}

\makeatletter
\def\@ssect#1#2#3#4#5#6{%
  \NR@gettitle{#6}
  \old@ssect{#1}{#2}{#3}{#4}{#5}{#6}
}
\makeatother

\begin{document}
\begin{frontmatter}

\title{Communication-Efficient Learning for Satellite Constellations\thanksref{footnoteinfo}} 

\thanks[footnoteinfo]{Corresponding author: N. Bastianello (\texttt{nicolba@kth.se}) \\
R.-S.T. and M.G. contributed equally.\\
The work of R.-S.T. and M.G. was supported by the Digital Futures Summer Research Internship Program.\\
The work of N.B. and K.H.J. was partially supported by the EU Horizon Research and Innovation Actions program under Grant 101070162, and by Swedish Research Council Distinguished Professor Grant 2017-01078 Knut and Alice Wallenberg Foundation Wallenberg Scholar Grant.}

\author[First]{Ruxandra-Stefania Tudose} 
\author[First]{Moritz H.W. Grüss} 
\author[Second]{Grace Ra Kim}
\author[Third]{Karl H. Johansson}
\author[Third]{Nicola Bastianello}

\address[First]{KTH Royal Institute of Technology, Sweden}
\address[Second]{Stanford University, United States of America}
\address[Third]{School of Electrical Engineering and Computer Science, and Digital Futures, KTH Royal Institute of Technology, Sweden}

\begin{abstract}
Satellite constellations in low-Earth orbit are now widespread, enabling positioning, Earth imaging, and communications.
In this paper we address the solution of learning problems using these satellite constellations. In particular, we focus on a federated approach, where satellites collect and locally process data, with the ground station aggregating local models.
We focus on designing a novel, communication-efficient algorithm that still yields accurate trained models. To this end, we employ several mechanisms to reduce the number of communications with the ground station (local training) and their size (compression). We then propose an error feedback mechanism that enhances accuracy, which yields, as a byproduct, an algorithm-agnostic error feedback scheme that can be more broadly applied.
We analyze the convergence of the resulting algorithm, and compare it with the state of the art through simulations in a realistic space scenario, showcasing superior performance.
\end{abstract}


\end{frontmatter}

\section{Introduction}
Low-Earth orbit (LEO) satellites have proved a revolutionary technology for several applications, especially positioning, Earth imaging, communications \citep{celikbilek_survey_2022}.
In recent years, increasing numbers of LEO satellites have been placed in orbit, driven by the rise of large constellations \citep{EPRS2025}.
These satellite constellations then enable the collection of vast sets of data, \textit{e.g.} Earth images, which can be used to train accurate models of different phenomena. Examples include models that can be trained for disaster navigation \citep{barmpoutis_review_2020}, real-time earthquake prediction \citep{zhai_fedleo_2024}, disease spread \citep{franch-pardo_spatial_2020}, food security \citep{aragon_cubesats_2018} and climate change \citep{shukla2021enhancing,so_fedspace_2022,zhai_fedleo_2024}.

However, enabling learning on data collected by satellites poses significant practical challenges, mainly due to the disparity between data generation and the bandwidth available to transmit the data to ground stations (GSs) on Earth \citep{so_fedspace_2022}. Indeed, communications to the GS are limited by sparse communication windows and delays.

\begin{figure}[!ht]
    \centering    \includegraphics[scale=0.1, trim={0 6.5cm 0 0}, clip]{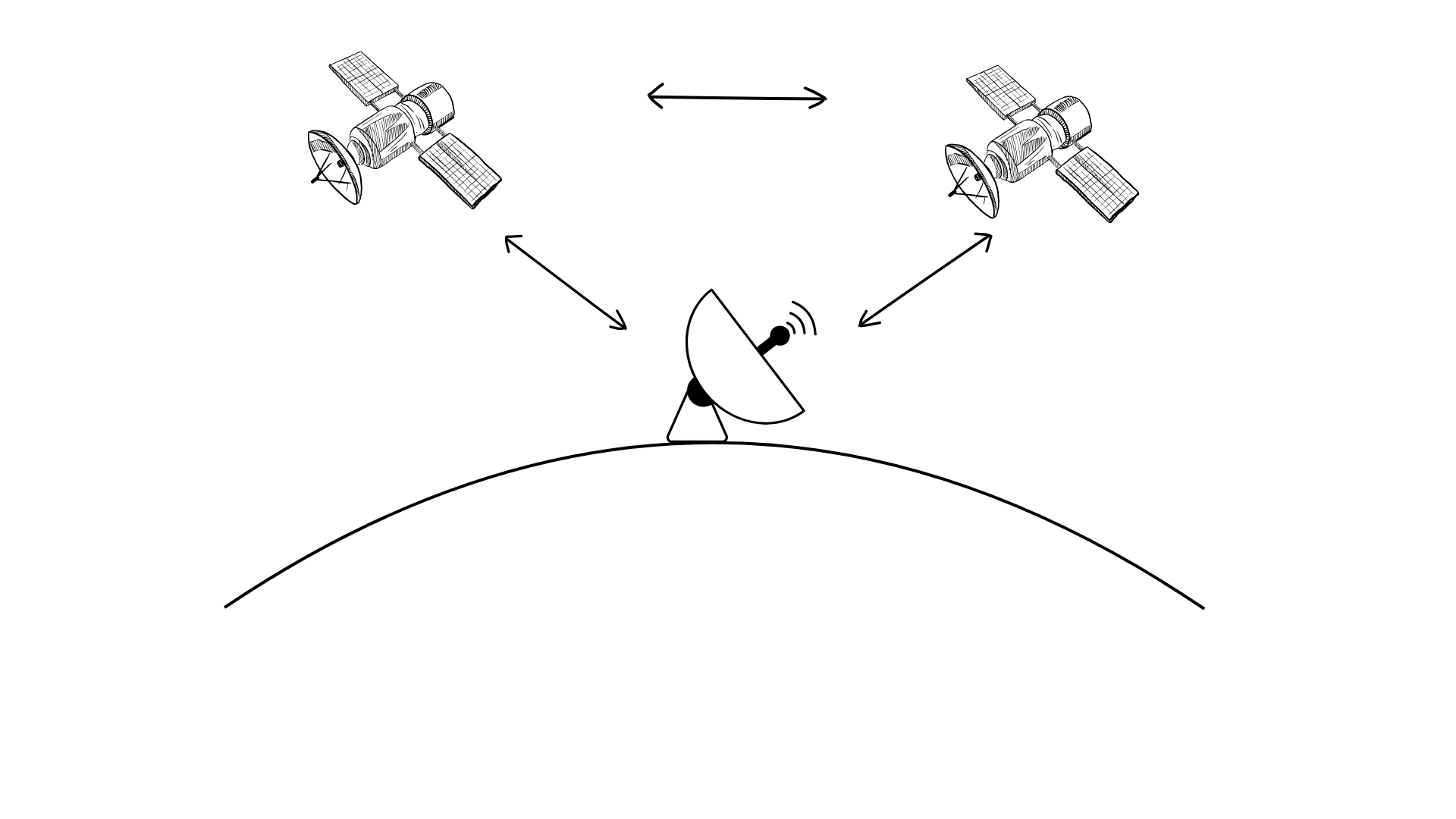}
    \caption{Federated learning for a satellite constellation, with ground station acting as coordinator.}
    \label{fig:fl-in-space}
\end{figure}

The solution that has been proposed to this issue is to leverage the paradigm of federated learning (FL) \citep{li_federated_2020}. The idea is to process data directly onboard the satellites, and to share only the resulting models. These models are transmitted to the GS that acts as a \textit{coordinator}, aggregating them and sending the result to the satellites for further training; see Figure~\ref{fig:fl-in-space} for a depiction of this scheme.
The resulting approach reduces significantly the bandwidth required (models, and not data, are transmitted) while still training accurate models on the data collected by all satellites in the constellation.
Apart from these notable features, FL brings another key contribution which is the enhanced security of the data. This is because in the FL architecture, the original data used for training never leaves the devices, only the model updates do \citep{uddin_sdn-based_2022}.

However, even when employing a federated learning approach, the bandwidth is still limited due to sparse communication windows to the GS.
In FL, several techniques have been proposed to enhance the communication efficiency of training algorithms. These can be categorized as compression/quantization (which reduces the size of messages) \citep{zhao_towards_2023}, partial participation (which reduces the number of agents transmitting to the coordinator) \citep{nemeth_snapshot_2022}, and local training (which reduces the frequency of agent-coordinator communications) \citep{grudzien_can_2023}.
In satellite applications, the use of both local training and partial participation have been tested, see \citep{so_fedspace_2022,zhai_fedleo_2024,razmi_scheduling_2022,han_cooperative_2024}. In particular, satellites perform local training while they do not have a communication window to the GS. Moreover, scheduling only a limited set of satellites to communicate with the GS leads to partial participation and hence reduction of the number of messages to be transmitted \citep {kim_bringing_2025}. 
Although these algorithms showcase practical improvements in performance, theoretical insights are still lacking. Additionally, communication compression was not employed, while in practice it is an important measure to improve communication efficiency. In this paper we aim to incorporate compression, and provide suitable theoretical results.

However, compression can reduce the accuracy of the trained model. The idea then is to employ error feedback (EF) to reduce compression's impact on accuracy \citep{karimireddy_error_2019}. As the name suggests, error feedback enacts a feedback on the compression error (that is, the difference between compressed and uncompressed messages), so that this error is integrated in future messages, preventing loss of information. Integrating EF thus allows to improve the accuracy of the trained model while still having communication efficiency.

In this paper we propose a satellite-ready FL algorithm designed to guarantee both communication efficiency and training accuracy. We start from the foundation of Fed-LT\footnote{We call Fed-LT the algorithm Fed-PLT of \citep{bastianello_enhancing_2024} without privacy (P) mechanism.} presented in \citep{bastianello_enhancing_2024}, which uses local training and partial participation to reduce the frequency and volume of communications. To these we add compression to then reduce the size of communications. To ensure accuracy, we also implement an error feedback mechanism similar to \citep{cheng_communication-efficient_2024}.
The main contributions are summarized as follows:
\begin{itemize}
    \item We design a novel FL algorithm that integrates a suite of communication efficiency tools (local training, partial participation, compression) while ensuring accuracy through error feedback. The compression is applied bi-directionally between agents and the coordinator.

    \item As a byproduct of this design, we present an algorithm-agnostic implementation of error feedback that can be plugged into any existing federated method.

    \item We further present a satellite-ready version of the algorithm, called Fed-LTSat, which integrates inter-satellite communications, which enhance accuracy while reducing the number of satellite-to-Earth communications.

    \item We test the resulting algorithm in numerical simulations implemented in FLySTacK, a platform that simulates space constraints \citep{kim_space_2024}, and compare with the state of the art.
\end{itemize}

\section{Algorithm Design and Convergence Analysis}
We start by formalizing the problem we aim to solve, then discuss the communication-efficient Fed-LT with EF, and its satellite-ready version. We conclude by analyzing their convergence.

\subsection{Problem formulation}
\begin{figure}[!ht]
    \centering
    \includegraphics[scale=0.15, trim={0 3.5cm 0 1cm}, clip]{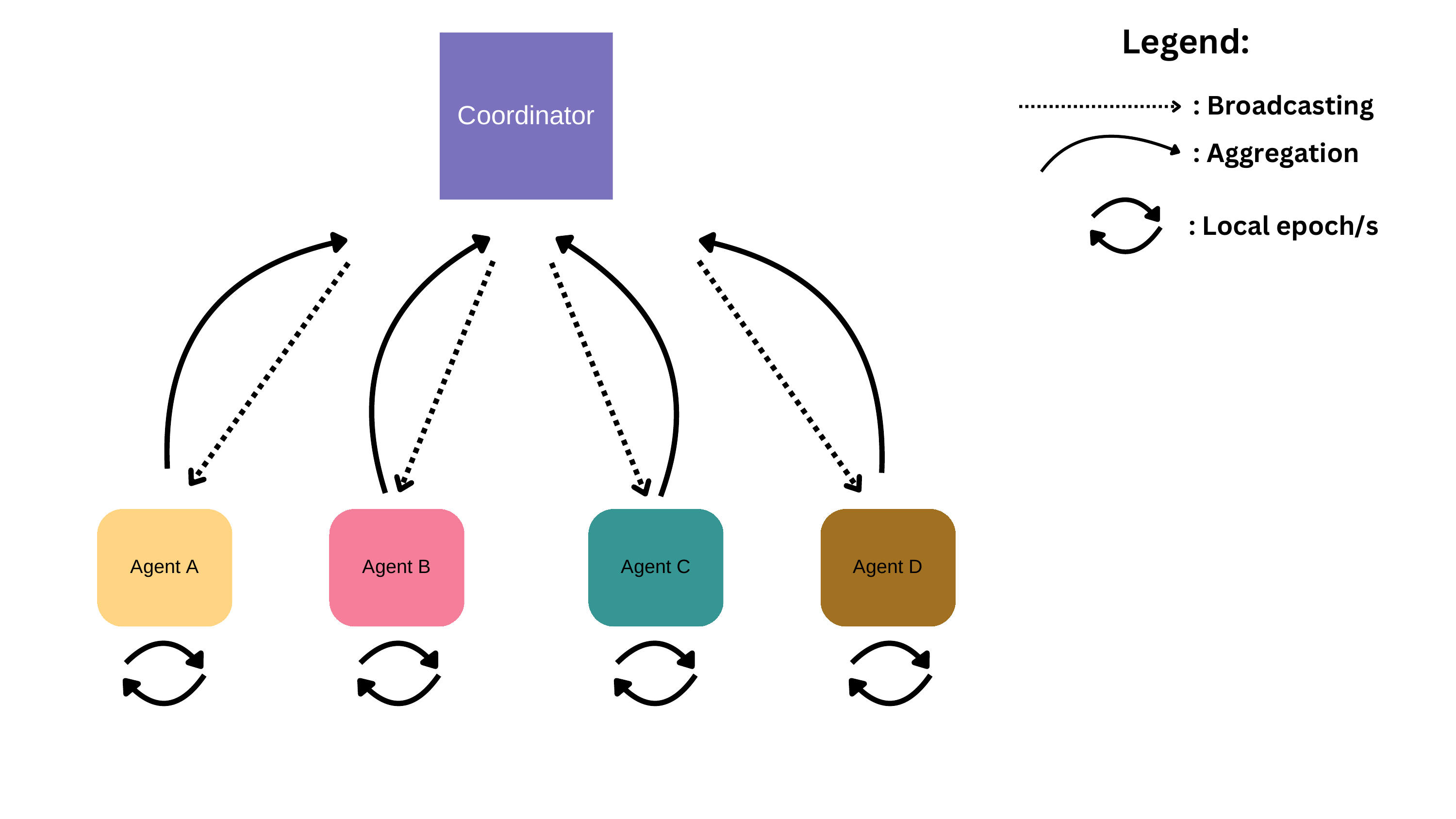}
    \caption{Scheme of the federated learning set-up.}
    \label{fig:fl-setup}
\end{figure}

The abstract federated learning set-up is depicted in Figure~\ref{fig:fl-setup}, with each of $N$ agents (\textit{e.g.} satellites) locally storing and processing data, and the coordinator (\textit{e.g.} ground station) aggregating the local models.
Formally, each agent stores a loss function $f_i : \mathbb{R}^n \to \mathbb{R}$, usually in an empirical risk form:
$$
    f_i(x) = \frac{1}{m_i} \sum_{h = 1}^{m_i} \ell(x, d_i^h)
$$
with $\{ d_i^h \}_{h = 1}^{m_i}$ the local data and $\ell : \mathbb{R}^n \times \mathbb{R}^p \to \mathbb{R}$ a loss function (\textit{e.g.} logistic loss).
These losses then define the following problem:
\begin{subequations}\label{eq:problem}
\begin{align}
    &\min_{x_i \in \mathbb{R}^n, \ i = 1, \ldots, N} \; \sum_{i=1}^N f_i(x_i) \\
    &\qquad \text{s.t.} \ x_1 = \ldots = x_N
\end{align}
\end{subequations}
which is iteratively solved through the scheme of Figure~\ref{fig:fl-setup}.

\subsection{Algorithm design: communication efficiency}
The foundation of our proposed algorithm is Fed-LT from \citep{bastianello_enhancing_2024}, which already incorporates local training and partial participation.
The goal of this section is to further enhance Fed-LT's communication efficiency by integrating compression, while still guaranteeing accuracy through error feedback.

In Algorithm~\ref{alg:algorithm-with-compression} we report Fed-LT \citep{bastianello_enhancing_2024} with the addition of compression on the uplink (agent-to-coordinator transmissions) and on the downlink (coordinator to agents broadcasts).
\begin{algorithm}[!ht]
\caption{Fed-LT with compression}
\label{alg:algorithm-with-compression}
\begin{algorithmic}[1]
	\Require For each agent initialize $x_{i,0}$ and $z_{i,0}$; choose the local solver, its parameters, and the number of local epochs $\Ne$; the parameter $\rho > 0$; choose the uplink $C_u$ and downlink $C_d$ compression; let $S_k \subset \{ 1, \ldots, N\}$ denote the agents active at iteration $k$
	\For{$k = 0, 1, 2, \ldots,$}
        \CommentState{coordinator update}
        \State receive $z_{i,k}$ from each agent $i \in S_k$
        \State aggregate
        $$
            y_{k+1} = \frac{1}{N} \left( \sum_{i \in S_k} z_{i,k} + \sum_{i \not\in S_k} z_{i,k-1} \right)
        $$
        \State compress \(y_{k+1}\),  \(C_d(y_{k+1})\), and transmits it to all active agents \Comment{downlink compression}

    	\CommentState{agents updates}
        \State gather the active agents in $S_{k+1}$
        \For{$i = 1, \ldots, N$}
            \If{$i \in S_{k+1}$} \Comment{partial participation}
                \State set $w_{i,k}^0 = x_{i,k}$, $v_{i,k} = 2 y_{k+1} - z_{i,k}$
                \For{$\ell = 0, 1, \ldots, \Ne-1$} \Comment{local training}
                    \State \textit{e.g.} gradient descent
                    $$
                        \hspace{1.45cm}w_{i,k}^{\ell+1} = w_{i,k}^\ell - \gamma \left( \nabla f_i(w_{i,k}^\ell) + \frac{1}{\rho} (w_{i,k}^\ell - v_{i,k})  \right)
                    $$
                \EndFor
                \State set $x_{i,k+1} = w_{i,k}^{\Ne}$
                \State update the auxiliary variable
                $$
                    z_{i,k+1} = z_{i,k} + 2(x_{i,k+1} - y_{k+1})
                $$
                \State compress $z_{i, k+1}$, \( C_u(z_{i,k+1})\), and transmit to coordinator \Comment{uplink compression}
            \Else \Comment{inactive agent}
                \State set $x_{i,k+1} = x_{i,k}$, $z_{i,k+1} = z_{i,k}$
            \EndIf
        \EndFor        
    \EndFor
\end{algorithmic}
\end{algorithm}
During the execution of Fed-LT, each (active) agent locally trains the model $x_i$ (using the local loss $f_i$ and a local solver like gradient descent), and updates the auxiliary variable $z_i$. The latter is compressed (via compression operator $C_u$) and transmitted to the coordinator.
The coordinator then aggregates the $z_i$'s received (using past information for inactive agents), compresses the result with $C_d$, and broadcasts it to all agents.

\begin{figure}[!ht]
    \centering
\includegraphics[width=\linewidth]{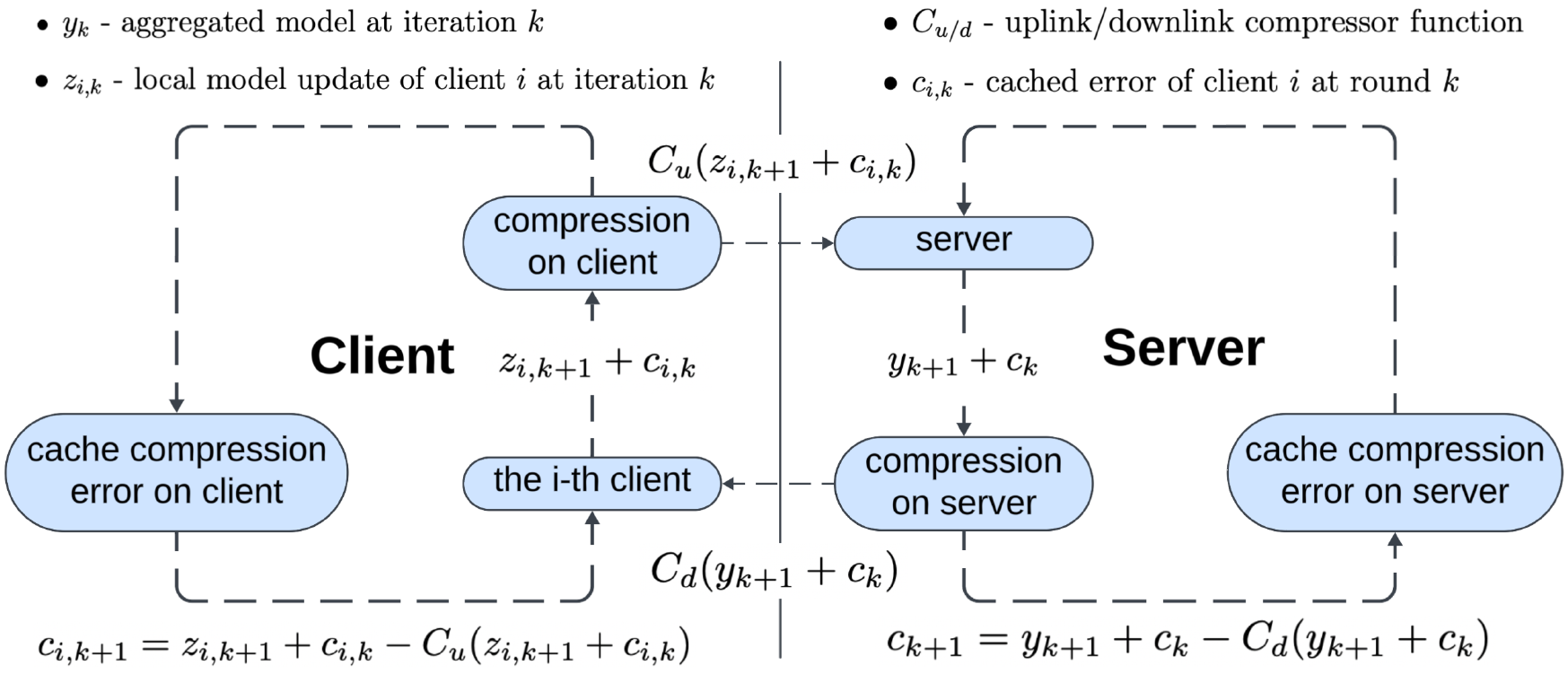}
    \caption{Flow chart of the agnostic error-feedback mechanism.}
    \label{fig:EF-flowchart}
\end{figure}

Clearly, the bi-directional compression in Algorithm~\ref{alg:algorithm-with-compression} causes loss of information twice, characterized by the compression errors $C_u(z_{i,k}) - z_{i,k}$ and $C_u(y_{k+1}) - y_{k+1}$.
The goal then is to integrate an error feedback mechanism inspired by \citep{cheng_communication-efficient_2024} to improve accuracy.
The idea is to cache the compression errors (both up- and downlink) to propagate them with later transmissions, reducing the loss of information. In particular, every time compression is applied, the error, \textit{e.g.} $c_{i,k} = C_u(z_{i,k}) - z_{i,k}$, is stored locally, and added to the message to be transmitted in the following iteration $z_{i,k+1} + c_{i,k}$. The resulting message is compressed, $C_u(z_{i,k+1} + c_{i,k})$, its compression error is cached and the message is transmitted.
The mechanism is applied to the up- and downlink at each iteration, ensuring that all information is ultimately transmitted.
The resulting EF scheme is depicted in Figure~\ref{fig:EF-flowchart}.
We remark that this scheme is actually \textit{algorithm-agnostic} and can be directly plugged into alternative federated algorithms, denoting by $z_{i,k}$ and $y_k$ the up- and downlink messages, respectively.
In particular, the application of Figure~\ref{fig:EF-flowchart} to Algorithm~\ref{alg:algorithm-with-compression} yields Algorithm~\ref{alg:main-algorithm with EF}.
\begin{algorithm}[!ht]
\caption{Fed-LT with compression and EF}
\label{alg:main-algorithm with EF}
\begin{algorithmic}[1]
	\Require For each agent initialize $x_{i,0}$ and $z_{i,0}$, and the cache $c_{i,0} = 0$; initialize the coordinator cache $c_0 = 0$; choose the local solver, its parameters, and the number of local epochs $\Ne$; the parameter $\rho > 0$; choose the uplink $C_u$ and downlink $C_d$ compression; let $S_k \subset \{ 1, \ldots, N\}$ denote the agents active at iteration $k$
	\For{$k = 0, 1, 2, \ldots,$}
        \CommentState{coordinator update}
        \State receive $z_{i,k}$ from each agent $i \in S_k$
        \State aggregate \Comment{downlink EF}
        $$
            y_{k+1} = c_k + \frac{1}{N} \left( \sum_{i \in S_k} z_{i,k} + \sum_{i \not\in S_k} z_{i,k-1} \right)
        $$
        \State compress \(y_{k+1}\),  \(C_d(y_{k+1})\), and transmits it to all active agents \Comment{downlink compression}
        \State cache the compression error $c_{k+1} = y_{k+1} - C_d(y_{k+1})$

    	\CommentState{agents updates}
        \State gather the active agents in $S_{k+1}$
        \For{$i = 1, \ldots, N$}
            \If{$i \in S_{k+1}$} \Comment{partial participation}
                \State set $w_{i,k}^0 = x_{i,k}$, $v_{i,k} = 2 y_{k+1} - z_{i,k}$
                \For{$\ell = 0, 1, \ldots, \Ne-1$} \Comment{local training}
                    \State \textit{e.g.} gradient descent
                    $$
                        \hspace{1.45cm}w_{i,k}^{\ell+1} = w_{i,k}^\ell - \gamma \left( \nabla f_i(w_{i,k}^\ell) + \frac{1}{\rho} (w_{i,k}^\ell - v_{i,k})  \right)
                    $$
                \EndFor
                \State set $x_{i,k+1} = w_{i,k}^{\Ne}$
                \State update the auxiliary variable
                $$
                    z_{i,k+1} = z_{i,k} + 2(x_{i,k+1} - y_{k+1})
                $$
                \State compress $z_{i, k+1} + c_{i,k}$, \( C_u(z_{i,k+1} + c_{i,k})\), and transmit to coordinator \Comment{uplink EF}
                \State cache the compression error
                $$
                    c_{i,k+1} = z_{i, k+1} + c_{i,k} - C_u(z_{i,k+1} + c_{i,k})
                $$
            \Else \Comment{inactive agent}
                \State set $x_{i,k+1} = x_{i,k}$, $z_{i,k+1} = z_{i,k}$, $c_{i,k+1} = c_{i,k}$
            \EndIf
        \EndFor        
    \EndFor
\end{algorithmic}
\end{algorithm}

\begin{remark}
The Fed-LT framework allows for customization of the local training, by selecting \textit{e.g.} (stochastic) gradient descent, noisy gradient descent for privacy preservation \citep{bastianello_enhancing_2024}. Additionally, the aggregation step performed by the coordinator can be robustified against model attacks \citep{pasdar_robust_2025}.
\end{remark}

\subsection{Algorithm design: satellite-ready version}
Algorithm~\ref{alg:main-algorithm with EF} in principle could be implemented over a satellite constellation to enable communication-efficient learning. However, even when employing compression the satellite-to-ground station communications still represent the main bottleneck, \textit{e.g.} due to communication windows and delays. The idea then is to further modify Algorithm~\ref{alg:main-algorithm with EF} to enhance its efficiency.

Surveying the literature on FL for satellite constellations, two main approaches stand out: 1) the ground station serves as coordinator \citep{so_fedspace_2022}, 2) aggregation is performed via Inter-Satellite Links (ISLs) \citep{zhai_fedleo_2024,kim_bringing_2025}. The drawback of 1) is that the ground station is used continuously throughout training, requiring long-range communications, while 2) employs communications between satellites in closer range. However, 1) yields higher accuracy of the trained model.
The idea then is to modify Algorithm~\ref{alg:main-algorithm with EF} to employ both modes of communication, satellite-to-ground station and inter-satellite, balancing communication efficiency and accuracy. In particular, instead of requiring several satellites to directly communicate with the GS, we allow some satellites to receive the local updates of neighboring satellites in the same orbit and forward them to the GS. This way fewer satellite-to-ground station links need to be established, while still allowing the participation of several satellites.
An additional augmentation to ``space-ify" \citep{kim_bringing_2025}. Algorithm~\ref{alg:main-algorithm with EF} comes in the form of \textit{satellite-ready partial participation}. The idea is to select which satellites will connect to the GS and transmit updates based on the analysis of their orbits, in order to minimize the total time to execute a round of communication. The selected satellites can then serve to forward the model updates of neighboring satellites according to the mechanism described above.

The space-ified version of Algorithm~\ref{alg:main-algorithm with EF} is reported in Algorithm~\ref{alg:main-algorithm}, and is characterized by the features summarized as follows:
\begin{itemize}
    \item the satellites collect and locally store data (\textit{e.g.} Earth images);

    \item this data is processed onboard the satellites, so only trained models, and not large datasets, are communicated to Earth (lines 7-13);

    \item the schedule of satellite-to-GS communications is optimized to the time of a communication round (line 6);

    \item and in addition, satellites connected to the GS also serve to forward the models received through inter-satellite links (line 15).
\end{itemize}

\begin{algorithm}[!ht]
\caption{Space-ified Fed-LT (Fed-LTSat)}
\label{alg:main-algorithm}
\begin{algorithmic}[1]
	\Require For each agent initialize $x_{i,0}$ and $z_{i,0}$, and the cache $c_{i,0} = 0$; initialize the coordinator cache $c_0 = 0$; choose the local solver, its parameters, and the number of local epochs $\Ne$; the parameter $\rho > 0$; choose the uplink $C_u$ and downlink $C_d$ compression; let $S_k \subset \{ 1, \ldots, N\}$ denote the agents active at iteration $k$
	\For{$k = 0, 1, 2, \ldots,$}
        \CommentState{coordinator update}
        \State receive $z_{i,k}$ from each agent $i \in S_k$
        \State aggregate \Comment{downlink EF}
        $$
            y_{k+1} = c_k + \frac{1}{N} \left( \sum_{i \in S_k} z_{i,k} + \sum_{i \not\in S_k} z_{i,k-1} \right)
        $$
        \State compress \(y_{k+1}\),  \(C_d(y_{k+1})\), and transmits it to all active agents \Comment{downlink compression}
        \State cache the compression error $c_{k+1} = y_{k+1} - C_d(y_{k+1})$

    	\CommentState{agents updates}
        \State select the active agents $S_{k+1}$ using the scheduler of \citep{kim_bringing_2025} ($S_{k+1}$ includes satellites directly connected to the GS, and satellites connected to neighboring ones that forward their updates) \Comment{space-ification}
        \For{$i = 1, \ldots, N$}
            \If{$i \in S_{k+1}$} \Comment{partial participation}
                \State set $w_{i,k}^0 = x_{i,k}$, $v_{i,k} = 2 y_{k+1} - z_{i,k}$
                \For{$\ell = 0, 1, \ldots, \Ne-1$} \Comment{local training}
                    \State \textit{e.g.} gradient descent
                    $$
                        \hspace{1.45cm}w_{i,k}^{\ell+1} = w_{i,k}^\ell - \gamma \left( \nabla f_i(w_{i,k}^\ell) + \frac{1}{\rho} (w_{i,k}^\ell - v_{i,k})  \right)
                    $$
                \EndFor
                \State set $x_{i,k+1} = w_{i,k}^{\Ne}$
                \State update the auxiliary variable
                $$
                    z_{i,k+1} = z_{i,k} + 2(x_{i,k+1} - y_{k+1})
                $$
                \State compress $z_{i, k+1} + c_{i,k}$, \( C_u(z_{i,k+1} + c_{i,k})\), and transmit to the GS or to a neighboring satellite for forwarding \Comment{uplink EF}
                \State cache the compression error
                $$
                    c_{i,k+1} = z_{i, k+1} + c_{i,k} - C_u(z_{i,k+1} + c_{i,k})
                $$

            \Else \Comment{inactive agent}
                \State set $x_{i,k+1} = x_{i,k}$, $z_{i,k+1} = z_{i,k}$, $c_{i,k+1} = c_{i,k}$
            \EndIf
        \EndFor
    \EndFor
\end{algorithmic}
\end{algorithm}

\subsection{Convergence analysis}
In this section we analyze the convergence of the proposed Algorithm~\ref{alg:main-algorithm with EF}, which in turn establishes that of the space-ified version Algorithm~\ref{alg:main-algorithm}. We make the following assumption.

\begin{assumption}[Loss functions]\label{as:loss}
The local loss functions in~\eqref{eq:problem} are $\bar{\lambda}$-smooth and $\underline{\lambda}$-strongly convex.
\end{assumption}

This assumption ensures that~\eqref{eq:problem} admits a unique solution; future work will look at convergence for non-convex problems.
We introduce now the following definition of $\delta$-approximate compressor \citep{karimireddy_error_2019}, which we apply to the uplink and downlink compression in Assumption~\ref{as:compressor}.

\begin{definition}[$\delta$-approximate compressor]\label{def:compressor}
An operator $C : \mathbb{R}^n \to \mathbb{R}^n$ is a $\delta$-approximate compressor if there exists $\delta \in (0, 1]$ such that
$$
    \norm{C(x) - x}^2 \leq (1 - \delta) \norm{x}^2, \quad \forall x \in \mathbb{R}^n.
$$
\end{definition}

\smallskip

\begin{assumption}[Compressors]\label{as:compressor}
The uplink and downlink compressors, $C_u$ and $C_d$, are $\delta$-approximate.
\end{assumption}

We are now ready to state the following convergence result, proved in Appendix~\ref{app:proof}.

\begin{proposition}\label{pr:convergence}
Let Assumptions~\ref{as:loss} and~\ref{as:compressor} hold. Assume that each agent $i$ is active at iteration $k$ with probability $p_i \in (0, 1]$. Assume that $\norm{x_{i,k}} \leq \beta$. Then 
$$
    \lim_{k \to \infty} \norm{x_{i,k} - \bar{x}} \leq 2 \sqrt{\frac{\max_j p_j}{\min_j p_j}} \frac{1}{1 - \sigma} \frac{\sqrt{1 - \delta}}{\delta} \beta
$$
where $\bar{x} = \argmin_{x \in \mathbb{R}^n} \sum_{i = 1}^N f_i(x)$, $\sigma \in (0, 1)$ is the convergence rate characterized in \citep[Proposition~2]{bastianello_enhancing_2024}.
\end{proposition}

\section{Numerical Results}
In this section we evaluate the proposed algorithms applied to a classification task.
We start in section~\ref{subsec:numerical-ef} by comparing Algorithm~\ref{alg:algorithm-with-compression} and Algorithm~\ref{alg:main-algorithm with EF} to evaluate the impact of error feedback. Then in section~\ref{subsec:numerical-space} we compare Fed-LTSat (Algorithm~\ref{alg:main-algorithm}) with state-of-the-art alternatives in the realistic space simulator FLySTacK \citep{kim_space_2024}.

The classification task to be solved throughout this section is problem~\eqref{eq:problem} with local losses characterized by
\begin{equation}\label{eq:local loss function}
    f_i(x) = \frac{1}{m_i} \sum_{h=1}^{m_i} \log\left(1 + \exp(-b_{i,h} a_{i,h} x)\right) + \frac{\epsilon}{2N} \|x\|^2
\end{equation}
where $\epsilon = 50$, $m_i = 500$ for all agents, $n = 100$, and $N = 100$. The data are randomly generated.
We use $N_e = 10$, and all other hyperparameters are tuned optimally using grid search.

As a performance metric we employ the optimality error characterized by $e_k = \sum_{i = 1}^N \| x_{i,k} - \bar{x} \|^2$ where $\bar{x} = \argmin_{x \in \mathbb{R}^n} \sum_{i = 1}^N f_i(x)$. We also evaluate its asymptotic value $e_K$ with $K$ being the length of the simulation.

\subsection{Evaluating the impact of error feedback}\label{subsec:numerical-ef}
In this section we compare Fed-LT with bi-directional compression without EF (Algorithm~\ref{alg:algorithm-with-compression}) and with EF (Algorithm~\ref{alg:main-algorithm with EF}). For simplicity, we do not employ partial participation.
The compression applied on both uplink and downlink is characterized by Definition~\ref{quantization}.

\begin{definition}[Uniform quantization]\label{quantization}
Let $x \in \mathbb{R}^n$ and $[x]_i$ denote its $i$-th component. The uniform quantization applies the function $q : \mathbb{R} \to \mathbb{R}$ component-wise:
\begin{equation}
    q(x) = \Delta \cdot \left\lfloor \frac{x - V_{\min}}{\Delta} + 0.5 \right\rfloor + V_{\min}
\label{eq:quant}
\end{equation}
where \(
\Delta = \frac{V_{\max} - V_{\min}}{L}
\) is the step size, \( V_{\max} \) and \( V_{\min} \) the maximum and minimum values, and \( L \) the number of quantization levels. The result is the vector $( q([x]_1), \ldots, q([x]_n))$.
\end{definition}

We average the error across $20$ Monte Carlo simulations, each with $K = 500$ iterations, employing different values of $L$, $V_{\min}$ and $V_{\max}$.
\begin{table}[!ht]
\centering
\caption{Comparison with and without error feedback}
\label{tab:fedplt-sims}
\small
\renewcommand{\arraystretch}{1.2}
\setlength{\tabcolsep}{4pt}

\definecolor{headergray}{gray}{0.9}

\begin{tabular}{c|c|c|c}
\hline
\rowcolor{headergray}
 & $L$ & $V_{\min}, V_{\max}$ & \shortstack{Asymptotic\\Error} \\
\hline
Algorithm \ref{alg:algorithm-with-compression} & 1000 & $-10, 10$ & 0.01192 \\
                   & 10   & $-1, 1$  & 1.29873 \\
\hline
Algorithm \ref{alg:main-algorithm with EF} & 1000 & $-10, 10$ & 0.00348 \\
                           & 10   & $-1, 1$  & 0.37752 \\
\hline
\end{tabular}
\vspace{0.5em}
\end{table}
The results in terms of asymptotic error are reported in Table~\ref{tab:fedplt-sims}. We see that employing error feedback yields a significant performance improvement. We also remark that a more coarse quantization (smaller number of levels $L$) yields larger asymptotic optimality error, as more information is lost; although of course the communication size is smaller.

\begin{figure}[!ht]
    \centering
    \includegraphics[width=0.8\linewidth]{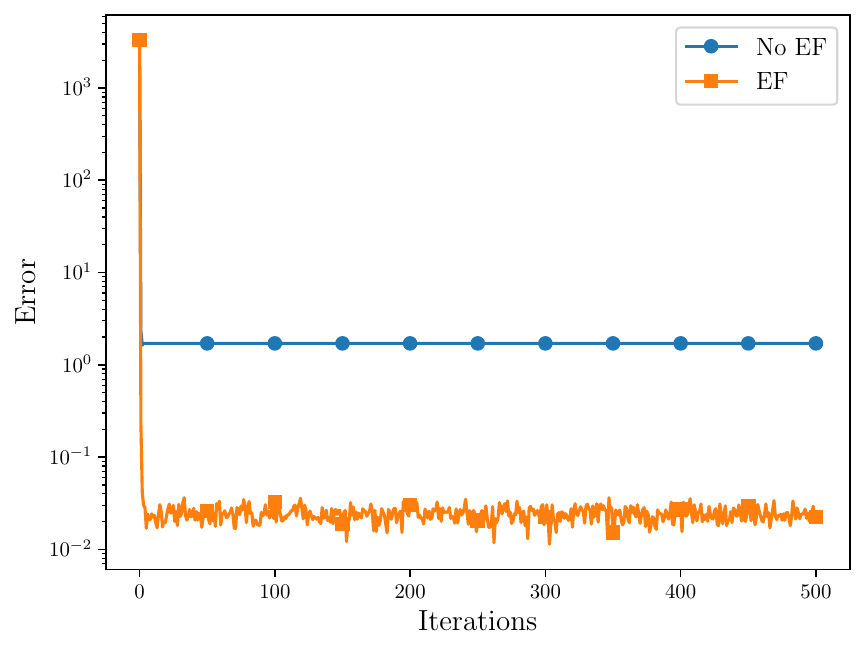}
    \caption{Optimality error evolution over iterations, employing quantization with $V_{\min} = - 1$, $V_{\max} =  1$ and $L=10$}
    \label{EFgraph}
\end{figure}

We conclude by presenting in Figure~\ref{EFgraph} an example of the optimality error evolution over the iterations of each algorithm. This illustrates the improvement in asymptotic optimality error brought about by using EF.

\begin{table*}[!ht]
\centering

\caption{Mean and standard deviation of the asymptotic optimality errors for the different algorithms, using $4$ compressors and with $10\%$ participation}

\resizebox{0.9\textwidth}{!}{%
\renewcommand{\arraystretch}{1.4}
\setlength{\tabcolsep}{6pt}
\definecolor{headergray}{gray}{0.9}

\begin{tabular}{l|c|c|c|c}
\hline
\rowcolor{headergray}
Algorithm &
\shortstack{Quantization\\($L=1000,\ V_{\min}= -10, \ V_{\max}= 10$)} &
\shortstack{Quantization\\($L=10,\ V_{\min}= -1, \ V_{\max}= 1$)} &
\shortstack{Rand-$d$\\($d=0.8 n$)} &
\shortstack{Rand-$d$\\($d=0.2 n$)} \\
\hline
Fed-LTSat (this paper) & $\mathbf{5.91 \times 10^{-5}\,(\pm 0.00 \times 10^{0})}$ & $\mathbf{3.11 \times 10^{-3}\,(\pm 2.77 \times 10^{-4})}$ & $\mathbf{7.85 \times 10^{-3}\,(\pm 2.15 \times 10^{-3})}$ & $1.14 \times 10^{0}\,(\pm 4.46 \times 10^{-1})$ \\
\hline
FedAvg  \citep{mcmahan_communication-efficient_2023} & $4.67 \times 10^{-4}\,(\pm 7.39 \times 10^{-5})$ & $7.64 \times 10^{0}\,(\pm 9.48 \times 10^{-2})$ & $1.06 \times 10^{0}\,(\pm 6.64 \times 10^{-2})$ & $3.06 \times 10^{0}\,(\pm 1.47 \times 10^{-1})$ \\
\hline
FedProx \citep{fedprox} & $2.74 \times 10^{-4}\,(\pm 1.97 \times 10^{-5})$ & $1.32 \times 10^{0}\,(\pm 2.94 \times 10^{-2})$ & $1.40 \times 10^{-2}\,(\pm 6.45 \times 10^{-3})$ & $1.03 \times 10^{0}\,(\pm 3.81 \times 10^{-2})$ \\
\hline
LED \citep{alghunaim_local_2024}    & $1.63 \times 10^{-2}\,(\pm 1.38 \times 10^{-4})$ & $1.57 \times 10^{0}\,(\pm 0.00 \times 10^{0})$ & $4.51 \times 10^{-1}\,(\pm 1.88 \times 10^{-1})$ & $9.02 \times 10^{-1}\,(\pm 3.16 \times 10^{-5})$ \\
\hline
5GCS \citep{grudzien_can_2023}   & $4.09 \times 10^{-2}\,(\pm 0.00 \times 10^{0})$ & $3.89 \times 10^{0}\,(\pm 0.00 \times 10^{0})$ & $9.66 \times 10^{-3}\,(\pm 4.65 \times 10^{-3})$ & $\mathbf{8.74 \times 10^{-1}\,(\pm 3.63 \times 10^{-1})}$ \\
\hline
\end{tabular}
}
\vspace{0.8em}

\label{tab:flystack-sims}
\end{table*}

\subsection{Comparison in a realistic space scenario}\label{subsec:numerical-space}
In this section we test Fed-LTSat (Algorithm~\ref{alg:main-algorithm}) in the realistic space scenario of FLySTacK \citep{kim_space_2024}. FLySTacK simulates a satellite constellation in LEO, determining when each satellite has a communication window to the ground station (which acts as coordinator). This information is then used to schedule which satellites should send an updated model, and which should do so through inter-satellites links.
In particular, we compare Fed-LTSat against the state-of-the-art alternatives FedAvg \citep{mcmahan_communication-efficient_2023}, FedProx \citep{fedprox}, LED \citep{alghunaim_local_2024}, and 5GCS \citep{grudzien_can_2023}.
These algorithms, however, were not designed for space applications, hence we space-ified them according to \citep{kim_bringing_2025}, further adding compression on the uplink and downlink communications. For a fair comparison, we apply the algorithm-agnostic error feedback scheme proposed in Figure~\ref{fig:EF-flowchart}, highlighting its flexibility.

We run $5$ Monte Carlo simulations in the space set-up, allowing $10$ satellites to participate at each round ($10\%$). The compression applied to uplink and downlink communications is either the quantization of Definition~\ref{quantization} or the rand-$d$ sparsification of Definition~\ref{Def:rand-k}.

\begin{definition}[Rand-$d$ sparsification]\label{Def:rand-k}
Let $x \in \mathbb{R}^n$ and $[x]_i$ denote its $i$-th component. The rand-$d$ sparsification is defined as follows: select uniformly at random the indices $\mathcal{I} \subset \{ 1, \ldots, n \}$, $|\mathcal{I}| = d$. Then return the vector $(\text{rand}([x]_1), \ldots, \text{rand}([x]_n))$ where
\[
    \text{rand}([x]_i) =
    \begin{cases}
    [x]_i & \text{if} \ i \in \mathcal{I}, \\
    0 & \text{otherwise}.
    \end{cases}
\]
\end{definition}

The results of the simulations are reported in Table~\ref{tab:flystack-sims}.
We can see that Fed-LTSat outperforms the alternative algorithms in most scenarios, and otherwise performs very close to the alternatives. In some scenarios (\textit{e.g.} quantization) the performance of Fed-LTSat is orders of magnitude better. When using quantization with $L = 10, V_{min}=-1, V_{max}=1$ for example, the asymptotic optimality error, is $3$ orders of magnitude lower, than any other simulated algorithm.
We also remark that, similarly to the results in section~\ref{subsec:numerical-ef}, a coarser quantization/sparsification yields larger asymptotic optimality errors, due to more information being omitted from communications.

\section{Conclusions and Future Work}
In this paper, we address the solution of learning problems with satellite constellations. In particular, we focus on designing a novel algorithm that ensures communication efficiency while preserving the accuracy of the trained model. We also show how to ``space-ify" this algorithm for deployment in satellite constellations. We provide numerical results showcasing the performance of the proposed algorithm and comparing it with the state of the art.
Future research directions include providing a broader theoretical framework for the algorithm-agnostic error feedback mechanism we proposed; and applying the proposed algorithm to non-convex problems.

\appendix
\section{Proof of Proposition~1}\label{app:proof}
By \citep{bastianello_enhancing_2024}, we can interpret Algorithm~\ref{alg:main-algorithm with EF} as the stochastic update
\begin{equation}
    \begin{bmatrix} x_{i, k+1} \\ z_{i,k+1} \end{bmatrix} = \begin{cases}
        \begin{bmatrix} \mathcal{X}_i(x_{i, k}, z_{i,k}) \\ z_{i,k} + 2 ( \mathcal{X}_i(x_{i, k}, z_{i,k}) - y_{k+1} ) \end{bmatrix} + e_{i,k} \\ \hspace{5cm} \text{if} \ u_{i,k} = 1 \\
        \begin{bmatrix} x_{i, k} \\ z_{i,k} \end{bmatrix} \hspace{4cm} \text{if} \ u_{i,k} = 0
    \end{cases}
\end{equation}
where $x_{i,k+1} = \mathcal{X}_i(x_{i, k}, z_{i,k})$ represents the output of the local training, $u_{i,k} \sim \text{Ber}(p_i)$ is the r.v. determining whether agent $i$ is active, and $e_{i,k} \in \mathbb{R}^{2n}$ is the error due to compression after EF is applied.

By \citep[Proposition~2]{bastianello_enhancing_2024}, if $\mathbb{E}[\norm{e_{i,k}}] \leq \nu$, then we know that
\begin{equation}\label{eq:optimality-error}
    \mathbb{E}\left[ \norm{\begin{bmatrix} \x_{k} - \bar{\x} \\ \z_{k} - \bar{\z} \end{bmatrix}} \right] \leq \sqrt{\frac{\max_j p_j}{\min_j p_j}} \left( \sigma^k \norm{\begin{bmatrix} \x_{0} - \bar{\x} \\ \z_{0} - \bar{\z} \end{bmatrix}} + \frac{1 - \sigma^k}{1 - \sigma} \nu \right)
\end{equation}
where $\x$ and $\z$ are the vectors collecting $x_i$ and $z_i$, respectively, and $\bar{\x} = \boldsymbol{1} \otimes \bar{x}$ and $\bar{\z}$ is a suitable fixed point.
Therefore we need to prove that the compression error, after applying error feedback, is bounded in norm by $\nu$.

By Jensen's inequality we have $\mathbb{E}[\norm{e_{i,k}}] = \mathbb{E}[\sqrt{\norm{e_{i,k}}^2}] \leq \sqrt{\mathbb{E}[\norm{e_{i,k}}^2]}$ and we need to bound $\mathbb{E}[\norm{e_{i,k}}^2]$.
With a small modification of \citep[Lemma~3]{karimireddy_error_2019}, using the assumption that $\norm{x_{i,k}}^2 \leq \beta^2$ and that $C_u$, $C_d$ are $\delta$-approximate, yields
\begin{equation}\label{eq:ef-error-bound}
    \mathbb{E}[\norm{e_{i,k}}^2] \leq 4 \frac{1 - \delta}{\delta^2} \beta^2.
\end{equation}

Using~\eqref{eq:ef-error-bound} into~\eqref{eq:optimality-error} and taking the limit $k \to \infty$ yields
\begin{equation*}
    \lim_{k \to \infty} \mathbb{E}\left[ \norm{\begin{bmatrix} \x_{k} - \bar{\x} \\ \z_{k} - \bar{\z} \end{bmatrix}} \right] \leq \sqrt{\frac{\max_j p_j}{\min_j p_j}} \frac{1}{1 - \sigma} \frac{\sqrt{1 - \delta}}{\delta^2} \beta.
\end{equation*}
The thesis follows since $\mathbb{E}[\norm{x_{i,k} - \bar{x}}] \leq \mathbb{E}\left[ \norm{\begin{bmatrix} \x_{k} - \bar{\x} \\ \z_{k} - \bar{\z} \end{bmatrix}} \right]$. \qed

\bibliography{ifacconf}

\end{document}